\documentclass[review]{elsarticle}
\usepackage{amsmath,amsfonts,amssymb}
\usepackage{graphicx}
\usepackage{tocloft}
\usepackage{multirow,tabularx}
\usepackage{silence}
\usepackage[font=footnotesize,labelfont=bf]{caption}
\usepackage[font=footnotesize,labelfont=bf]{subcaption}

\journal{Journal of \LaTeX\ Templates}

\begin{document}

\begin{frontmatter}

\title{Attention-based Fully Gated CNN-BGRU for Russian Handwritten Text}


\author[mymainaddress]{Abdelrahman Abdallah}
\author[mysecondaryaddress]{Mohamed  Hamada }
\author[mythirdaddress]{Daniyar Nurseitov}

\address[mymainaddress]{MSc Machine Learning \& Data Science, Satbayev University,
 Almaty, Kazakhstan}
\address[mysecondaryaddress]{Associate Professor, IS Department International IT University, Almaty, Kazakhstan}
\address[mythirdaddress]{National Open Research Laboratory for Information and
Space Technologies at Satbayev University, Almaty, Kazakhstan}
\begin{abstract}
This research approaches the task of handwritten text with attention encoder-decoder networks that are trained on Kazakh and Russian language. We developed a novel deep neural network model based on Fully Gated CNN, supported by Multiple bidirectional GRU and Attention mechanisms to manipulate sophisticated features that achieve 0.045 Character Error Rate (CER), 0.192 Word Error Rate (WER) and 0.253 Sequence Error Rate (SER) for the first test dataset and 0.064 CER, 0.24 WER and 0.361 SER for the second test dataset. Also, we propose fully gated layers by taking the advantage of multiple the output feature from Tahn and input feature, this proposed work achieves better results and We experimented with our model on the Handwritten Kazakh \& Russian Database (HKR). Our research is the first work on the HKR dataset and demonstrates state-of-the-art results to most of the other existing models.
\end{abstract}

\begin{keyword}
Handwriting recognition \sep
Fully gated convolutional neural networks \sep
Bidirectional gated recurrent unit \sep
Deep learning

\end{keyword}

\end{frontmatter}


\section{Introduction}
Today, handwriting recognition is a crucial task. Providing solutions to this problem will facilitate the business processes automation for many companies. A clear example is a postal company, where the task of sorting a large volume of alchemistical and parcels is a complicated issue.

Handwriting recognition (HWR) or Handwritten Text Recognition (HTR) is the capacity of a machine to obtain and interpret intelligible handwriting information from sources such as paper documents, images, touchscreens, and other tools. The handwriting recognition program manages to encode, performs accurate character segmentation, and identifies the most probable words.
Offline Handwriting Text Recognition is the task of converting letters or words into the images to a digital text, the input is a variable two-dimensional image, and the output is a sequence of characters. It’s a great value to human-machine contact and it can support the automated processing of handwritten documents. Also, it considers a sub-task of Optical Character Recognition (OCR), which mainly focuses on extracting text from scanned documents and natural scene images. The recognition of Russian handwriting poses particular challenges and advantages and has been more recently addressed than the recognition of text in other languages.

Previous approaches to the Offline Handwriting Text Recognition use Hidden Markov Models (HMM) for transcription tasks \cite{BUNKE19951399,inproceedings}, extracting features from images using a sliding window and then predicting character labels with a Hidden Markov model (HMM)\cite{rabiner1986introduction,poritz1988hidden} is the prevalent automatic speech recognition approach \cite{bahl1983maximum,rabiner1989tutorial,lee1990speech}. The key benefit of HMMs represents in their probabilistic nature, suitability for noise-corrupted signals like speech or handwriting, also their computational foundations behind the existence of efficient algorithms to change the model parameters automatically and iteratively. The success of HMM led many researchers to expand it to handwriting recognition, describing each word picture as a series of remarks. Two approaches can be differentiated according to the way in which this representation is performed: implicit segmentation \cite{gillies1992cursive,caesar1993recognition,mohamed1996handwritten}, which leads to a speech such as the representation of the handwritten word image, and explicit segmentation which involves a segmentation algorithm to divide words into simple units such as letters \cite{chen1994off,gilloux1995strategies}.
Despite the mentioned  benefits of this approach, there are some limitations \cite{NIPS2007_3213,7814068} compared to the new models that are using an encoding-decoders network which combines a convolutional neural network (CNN) with a bidirectional recurrent neural network and with a connectionist temporal classification (CTC) output layer \cite{4531750,xu2015show}.
Inspired by the latest advances in machine translation \cite{bahdanau2014neural}, automated question answer \cite{abdallah2020automated}, image captioning \cite{huang2019attention}, sentimental text processing\cite{hamada2019sentimental},and speech recognition \cite{chorowski2015attention}, we believe that encoder-decoder models with attention mechanisms \cite{bluche2017scan,kang2018convolve} will become the new state-of-the-art for HTR tasks.

Attention-based methods have been used to help the networks to learn the correct features and focus on the right features as well as alignment between an image pixel and target characters \cite{cho2014learning}. Attention increases the capacity of the network to collect the most important information for every part of the output sequence. Furthermore, attention networks are able to model language structures in the output sequence instead of just mapping the input to the correct output \cite{chorowski2015attention}.\\In this research, our contribution is to present a novel attention-based fully gated convolutional recurrent neural network, trained in Kazakh and Russian dataset \cite{nurseitov2020hkr}, it will be processed as follows:
\begin{itemize}
    \item Handwritten samples (forms) of keywords in Kazakh and Russian (Areas, Cities , Village , etc.).
    \item Handwritten Kazakh and Russian alphabet in Cyrillic.
    \item Handwritten samples (Forms) of poems in Russian.
\end{itemize}
The following section investigates the related work on Offline Handwriting Text Recognition. Section three demonstrates the attention-based fully gated convolutional recurrent neural network. Section four and five provides experimental results and analysis through testing data obtained from Kazakh and Russian dataset, conclusions and remarks are given in Section 6.

\section{Related Work}

(Anshul Gupta , 2011)\cite{gupta2011offline} This paper performs a analysis of various feature based classification strategies for the recognition of offline handwritten characters. It proposes an optimal character recognition technique after the experimentation. The approach proposed involves segmentation of a handwritten word using heuristics and artificial intelligence.Three Fourier descriptor combinations are used in parallel as vectors of the features. Support vector machine (SVM) is used as the classifier. Using the lexicon to verify the validity of the predicted word, post processing is performed.It is found that the results obtained by using the proposed CR system are satisfying.

(Bianne Bernard. 2011)\cite{bianne2011dynamic} The purpose of this study is to create an effective system of word recognition resulting from the combination of three handwriting recognizers. The key component of this hybrid framework is an HMM-based recognizer that takes complex and contextual knowledge into consideration for better writing device modeling. A state-tying method based on decision tree clustering was implemented for the modeling of the contextual units. Decision trees are built according to a collection of expert questions about how characters are created.

(Theodore Bluche , 2017)\cite{bluche2017scan} They presented an attentive model for the identification of end to end handwriting. this model didn’t need to input paragraph segmentation. The model was inspired by the recently introduced differential models that focus on voice recognition, image captioning or translation. The key difference with a multidimensional LSTM network is the implementation of hidden and overt focus. Their main contribution to the identification of handwriting illustrated in automated transcription without prior line segmentation, which was imperative in the previous approaches. In addition, the machine can learn the order of reading and it can handle bidirectional scripts such as Arabic. They performed tests on the popular IAM database and announce promising results in near future to full paragraph transcription.

(Theodore Bluche , 2017 )\cite{bluche2017gated} In this article, they proposed a new neural network architecture for state-of-the-art handwriting recognition as an alternative to recurrent neural networks in multidimensional long-term memory (MD-LSTM). The model CNN and a bidirectional LSTM decoder TO predict sequences of characters. The aim of this research is to generate generic, multilingual, and reusable features with the convolutionary encoder, leveraging more data for learning transfer. The architecture is also motivated by the need for fast GPU training and the need for quick decoding on CPUs.

(Joan Puigcerver , 2017)\cite{puigcerver2017multidimensional} this research  fulfilled the  state-of-the-art offline handwritten text recognition rely extensively on long-term multidimensional memory networks. This implies that long-term, two-dimensional dependencies, theoretically represented by multidimensional recurrent layers, may not be necessary to achieve a good accuracy of recognition, at least in the lower layers of the architecture. An alternative model is explored in this study, which relies only on convolutional and one-dimensional recurrent layers that achieve better or comparable results than those of the current state-of-the-art architecture and run much faster. Furthermore, they found that using random distortions as synthetic data increases during training significantly improves our model 's precision.

(Zi-Rui Wang ,2020 )\cite{wang2020writer}In this study , a novel WCNN-PHMM architecture for offline handwritten Chinese text recognition is proposed to address two key issues: the large  vocabulary of Chinese characters and the diversity of writing styles.By combining parsimonious HMM based on state of unsupervised learning based on the writer's code,  this approach demonstrates its superiority to other state-of-the-art approaches based on both experimental results and analysis.
Our work is the first research paper in our HKR Dataset\cite{nurseitov2020hkr}, this dataset is first open-source in Russian and Kazakh handwritten datasets until now there is no dataset in theses language available to the researchers
 
\begin{figure}[bh]
\centerline{\includegraphics[width=1.3\textwidth,keepaspectratio]{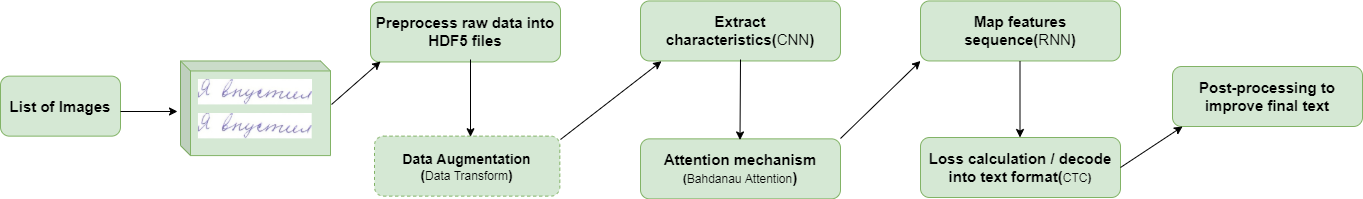}}

\caption{Atten-CNN-BGRU architecture Workflow.}
\end{figure}
\section{Proposed Model}

Our model focuses on the Cyrillic symbol extracted in handwritten form. An input is about as cropped word image of a 1D sequence of characters or symbols. Ours proposed a model based on Attention-Gated-CNN-BGRU architecture with few parameters (around 885,337), thus being high recognition rate, compact and faster, and low error rate compared with to other models.
The algorithm consists of six stages which will be described as follow:
\begin{enumerate}
    \item preprocessing such as  Resize with padding(1024x128), Illumination Compensation and Deslant Cursive Images then covert raw data into HDF5 files
    \item Extract characteristics by using CNN layers
    \item Bahdanau Attention mechanism that make the model to play attention to the inputs and related them to the output.
    \item Map features sequence by RRN 
    \item Loss calculation/decode into text format (CTC)
    \item Post-processing to improve the final text
\end{enumerate}
The workflow of our architecture shown in Fig. 1.
\subsection{Model}
In this section, we will describe our model, which the image go through a Gated CNN, then processes to Bahdanau attention, with bi-directional GRU, and finally the output matrix of GRU pass to the Connectionist Temporal Classification (CTC) \cite{graves2006connectionist} to calculate the loss value and also decode the output matrix  into the final text. Our model architecture shown in Fig. 2. which has four main parts encoder, attention, decoder, and CTC.
\begin{figure}[h!]
\centerline{\includegraphics[width=\textwidth,keepaspectratio]{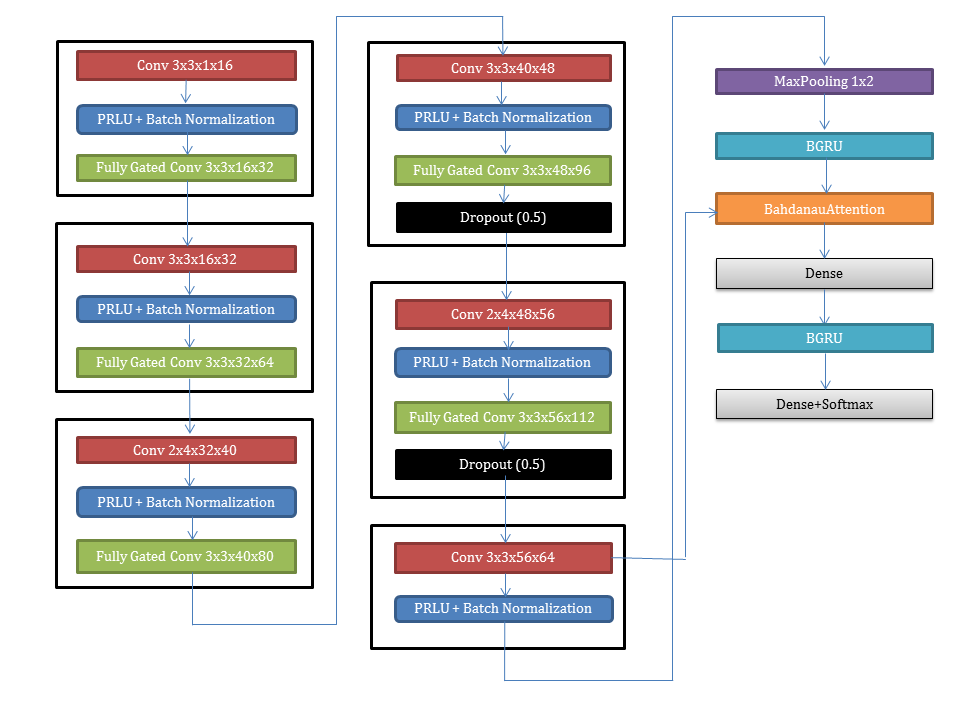}}

\caption{Atten-CNN-BGRU architecture for handwriting recognition.}
\end{figure}

\subsubsection{Encoder} 
\paragraph{Conventional blocks}
The encoder receives the input and generates the feature vectors. These feature vectors hold the information and the characteristics that represent the input. The encoder network consists of 5 convolutional blocks that correspond to training to extract relevant features from the images. Each block consists of convolution operation, which applies a filter kernel of size (3,3) in the first, second, fourth and sixth blocks and (2,4) in the third and fifth block, then Parametric ReLU and Batch Normalization are applied, also In order to reduce overfitting, we apply Dropout at some of convolutional layers \cite{srivastava2014dropout} (with dropout probability equal to 0.2).
\paragraph{Gated Conventional Layer}
The idea of gate controls is to propagate a feature vector to the next layer. The gate layer looks at the value of the vector feature at the given position, and at the adjacent values, and determines if it should be held or discarded at that position. It allows generic features to be computed across the entire image and filtered when the features are appropriate, depending on the context. 
The gate (g) layer is implemented as a convolutional layer with Tanh activation layer. It is added to the input function maps (x). The output of the gate mechanism is the point-wise multiplication of the inputs and outputs of the gate.

\begin{equation}
y=g(x).x
\end{equation}
where,
\begin{equation}
\begin{split}
g(x_{ij}) = tanh( \sum W X ) 
\end{split}
\end{equation}

In Fig. 3, showing a real example of what the output of the gate layer. In the examples, we display the feature maps of the output of the gated layer before and after. That example shows that the gated allows the feature to be effective and more excitatory

\begin{figure}
\begin{subfigure}{0.48\textwidth}
\includegraphics[width=\linewidth]{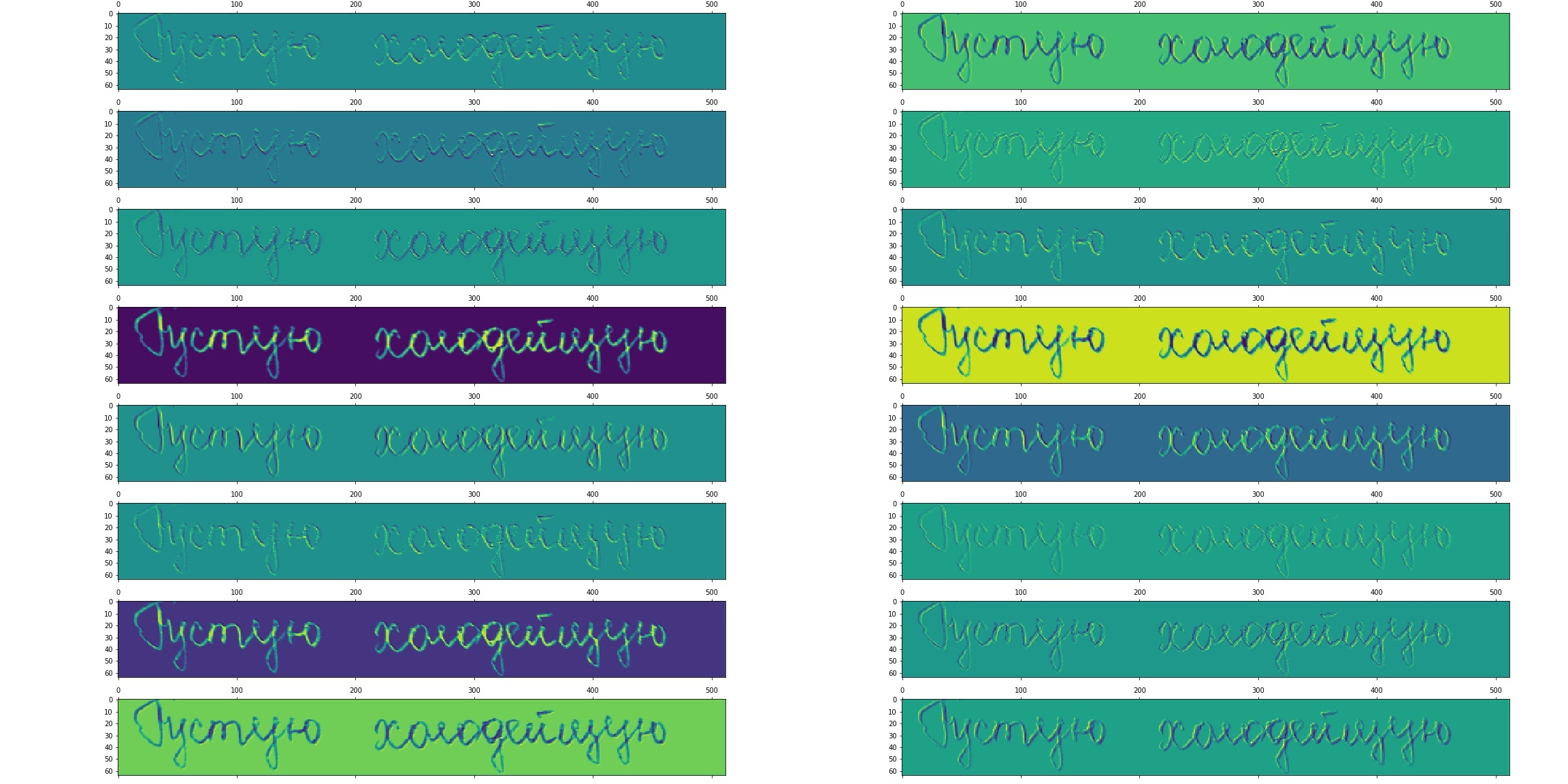}
\caption{Output features before gated layer}
\end{subfigure}
\hspace*{\fill}
\begin{subfigure}{0.48\textwidth}
\includegraphics[width=\linewidth]{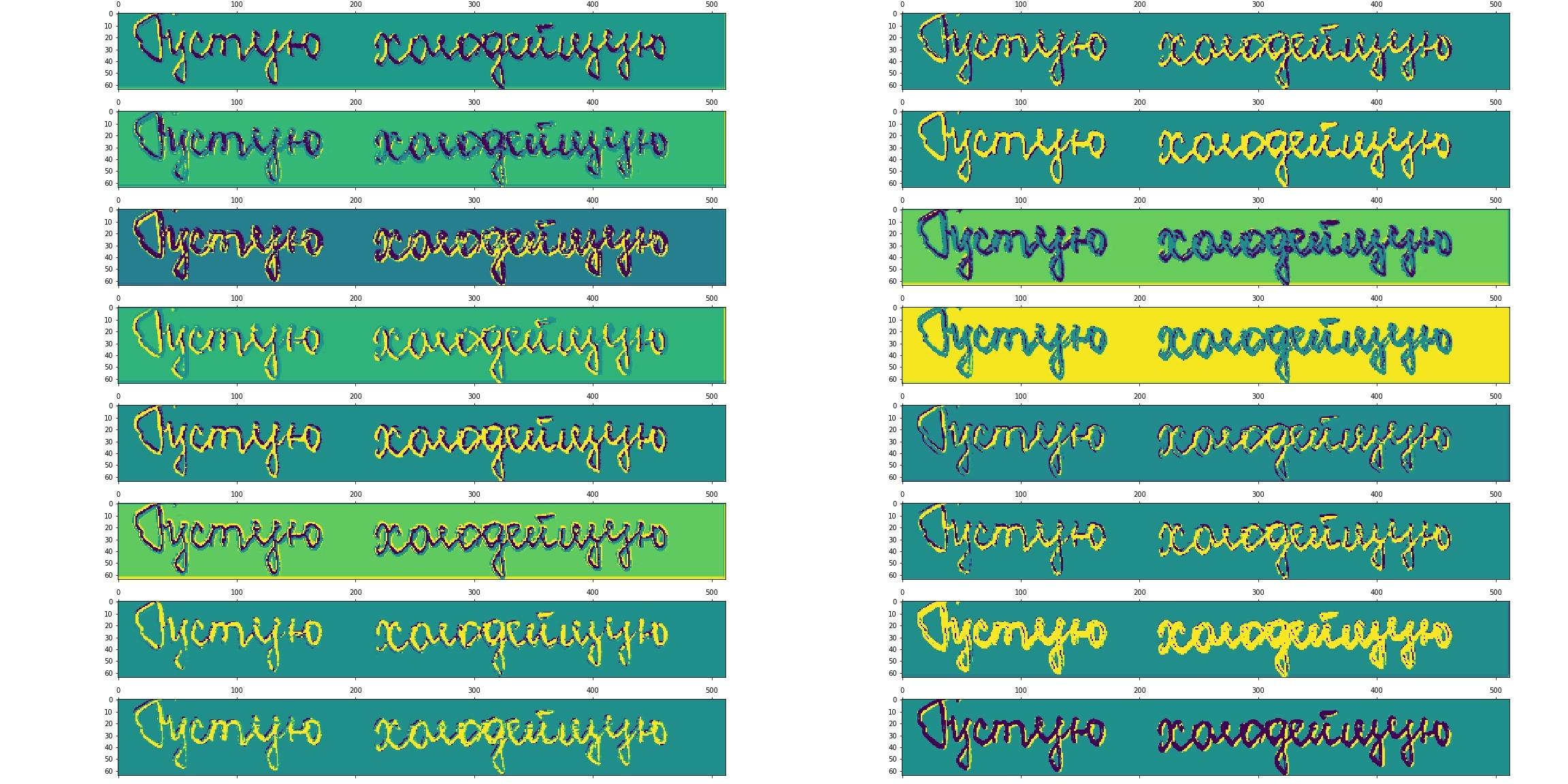}
\caption{Output features after gated layer}
\end{subfigure}
\caption{Visualization of the feature of a convolutional layer and  gate layer.} \label{A1E}
\end{figure} 

\subsubsection{Attention Mechanism} 
Attention is a mechanism   that   provides   a   richer   encoding   of   the   source sequence (h\textsubscript{1},....,h\textsubscript{s}) that facilitate the building of  a context vector (c\textsubscript{t}),  then the decoder can use it.

Attention allows the model to learn what are encoded images in the source to pay attention to, and to which degree during the prediction of each word in the target sequence. The hidden state of the source sequence is obtained from the encoder for each input time step, instead of the hidden state for the final time-step. 
\begin{equation}
\begin{split}
Attention\:weight(\alpha_{ts})=\frac{exp(score(h_{t},\Bar{h}_{s})}{\sum_{\Grave{s}}^{S} exp(score(h_{t},\Bar{h}_{\Grave{s}}))}
\end{split}
\end{equation}
\begin{equation}
\begin{split}
Context\:vector(c\textsubscript{t})=\sum_{s} \alpha_{ts} \Bar{h}_{s}
\end{split}
\end{equation}

In the target sequence, a context vector(c\textsubscript{t}) is constructed explicitly for every output word. Firstly, using a neural network, every hidden state from the encoder is graded  and  then  normalized  to  be  a  probability  over  the  hidden  states  of  the encoders. Finally,  the  probabilities  are  used  to  calculate  a  weighted  sum  of  the hidden states of the encoder to provide a context vector that should be used in the decoder. The attention layer produces outputs of dimension 128 x 256.

\begin{equation}
\begin{split}
Attention\:vector(a\textsubscript{t})=f(c\textsubscript{t},h\textsubscript{t})= tanh(W\textsubscript{c}[c_{t};h_{t}])
\end{split}
\end{equation}

\begin{equation}
\begin{split}
score(h_{t},\Bar{h}_{s})=\upsilon_{a}^{\top} tanh(W_{1}h_{t}+W_{1}\Bar{h}_{s})
\end{split}
\end{equation}
\subsubsection{Decoder}
The decoder is a bidirectional Gated Recurrent Unit (GRU) \cite{cho2014learning} RNN that processes feature sequences to predict sequences of characters. The feature vector contains 256 features per time-step, and the recurrent neural network propagates the information through this sequence. The GRU implementation of RNNs is employed, as it’s a gating mechanism in recurrent neural networks (RNN) almost like an extended LSTM unit without an output gate. GRU's trying to unravel the matter of the vanishing gradient.
A GRU's can solve vanishing gradients problem by using an update gate and a reset gate. The update gate can control the information that flows into the memory, and the reset gate can control the memory-flowing information. The gates for updating and resetting are two vectors that determine which information will be passed on to the output. They can be qualified to retain past knowledge or delete information unrelated to prediction. The GRU is similar to LSTM with a forgotten gate, but it contains fewer numbers of parameters because it lacks a gate for output. The output sequence of RNN layer is a matrix of size 128x96.

\subsubsection{Connectionist Temporal Classification(CTC)}
The connectionist temporal classification (CTC) output layer for recurrent neural networks, this for sequence labeling problems where there is no alignment between the inputs and the target labels. Neural networks need different training goals for each section of the input sequence or time-step.
This has two important consequences. First, it ensures that the training data must be pre-segmented in order to set the goals. Second, as the network generates only local classifications, the global aspects of the sequence (such as the probability of two labels occurring consecutively), must be modelled  externally. Indeed, the final label sequence can’t be inferred reliably without some sort of post-processing. This is achieved by allowing the network to make label predictions at any time in the input sequence, provided that the overall label sequence is correct. This can eliminate the necessity of pre-segmented data because it is no longer important to align the labels with the input. CTC also offers complete label sequence probabilities directly, which ensures that there is no additional post-processing is required for using the network as a time classifier.
While training the NN, the CTC is given the RNN output matrix and the ground truth text, and it computes the loss value. While inferring, the CTC is only given the matrix and it decodes it into the final text. Both the ground-truth text and the recognized text length can be mostly at 96 characters long.\\Loss Function: For a given input, we would like to train our model to maximize the probability which the correct answer is assigned to it.  To do this, we’ll need to calculate the conditional probability $p(Y \mid X)$. The function $p(Y \mid X)$ should also be differentiable so that we can be used gradient descent.\\For an alphabet A, a sequence labeling task in which the labels are drawn, CTC consists of a Softmax output layer with one or more units than the labels in A. Activations of the first $\mid A \mid$ units are the probabilities of output of the corresponding labels at specific times, given the input sequence and the network weights. The activation of the extra unit gives the probability that a 'blank' or no label will be output. The complete sequence of network outputs is then used to define the distribution over all possible label sequences of length up to the input sequence.\\Defining the alphabet $A^{`} = A \cup {blank} $,  The activation $y_{k}^{t}$ of network output k at time t is interpreted as the probability that the network will output element k of $A^{`}$ at time t, given the length of T input sequence x. Let $A^{`T}$ refer to the set of sequences T over $A^{`}$. Then, if we assume that the output probabilities are independent of those in other time-step at each time-step, we get the following conditional distribution over $ \pi\in A^{`T} $:

\begin{equation}
\begin{split}
p(\pi \mid x) = \prod_{t=1}^{T} y_{\pi t}^{t}	
\end{split}
\end{equation}
We now refer to the $\pi$ sequences over A as paths, to differentiate 
from sequences of marks, or marks L over A. The next stage consists of 
Defines the many-to-one function $F: A^{'T} \rightarrow A^{\leqq T}$, from the set of paths to the set $A^{\leqq T}$ of possible x labels. We do so by extracting the repeated labels from the paths first, and then the blanks. The probability of any $l \in A^{\leqq T}$ marking can be determined by summing the probabilities of all the paths described by F:
\footnotesize
\begin{equation}
\begin{split}
p(L \mid x) = \sum_{\pi \in F-1 } p(\pi | x)
\end{split}
\end{equation}

\section{Experiment Setup}
\subsection{Data}

The handwritten Kazakh, Russian database can serve as a basis for research on handwriting recognition. It contains Russian Words (Areas, Cities, Village, Settlements, Areas, Streets) by a hundred different writers. It also incorporates the most popular words in the Republic of Kazakhstan. A few preprocessing and segmentation procedures have been developed together with the database. It contains free handwriting forms in multiple areas of writer interest. This database is prepared for the purpose of providing a training and testing set for Kazakh, Russian Words recognition research.
This database consists of more than 1400 filled forms. There are approximately 63000 sentences, more than 715699  symbols, and there are approximately 106718 words.

\begin{figure}
\begin{subfigure}{0.48\textwidth}
\includegraphics[width=\linewidth]{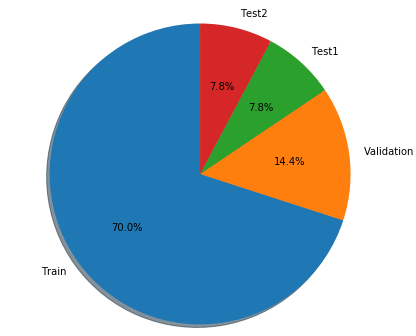}
\end{subfigure}
\hspace*{\fill}
\begin{subfigure}{0.48\textwidth}
\begin{tabular}{|c|c|} \hline
    Train images &  45470 \\ \hline
    Validation images & 9359 \\ \hline
    Test\_1 images &  5057 \\ \hline
    Test\_2 images & 5057  \\ \hline
    Total & 64943  \\ \hline
    \end{tabular}
\end{subfigure}
\caption{Training, validation and testing dataset} 
\end{figure} 

To train the research model, we used as much data as possible. Due to the scarcity of public data for Kazakh and Russian languages, we used our lab dataset which contains 64943 text lines and was divided as shown in Fig. 4. We evaluated our model in two different datasets, each dataset is separated carefully by the forms to ensure that it doesn’t include images of form in the other training and testing dataset.
The basis for this work’s dataset was made up of distinct BLABLA words (or short sentences) written in Russian and Kazakh languages (approximately   95\%  of Russian  and  5\%  of  Kazakh  words/sentences  respectively).  Considering that both of these languages  are  written  in  Cyrillic  and  share  the  same  33  characters.  Beside these characters, the Kazakh alphabet also contains 9 additional specific characters. This dataset of distinct BLABLA  words/sentences  were  boosted  by  applying  various handwriting styles ( approximately 50-100 different persons) to each of these distinct words. These procedures resulted in a final dataset with an overall number of BLABLA1 handwritten words/sentences are involved. Thereafter this final dataset was split into three datasets as follows: Training (70\%), Validation (15\%), and Testing (15\%). Test dataset itself was equally split into two sub-datasets (7.5\% each): the first dataset was named as TEST1 and consisted of words which do not exist in Training and Validation datasets; the second was  named  as  TEST2  and  made  up  of  words  that  exist  in  Training dataset  but  with  totally  different  handwriting  styles.  The main purpose of splitting the Test dataset into TEST1 and TEST2 was to check the accuracy of difference between recognition of unseen words and the other words which were seen in training phase but with unseen handwriting styles. After training, validation, and testing datasets had been prepared, and the models have been trained, a series of comparative evaluation experiments were conducted.
\subsection{Training}
We trained the our models to minimize the validation loss value of the the Connectionist Temporal Classification (CTC)  function. We performed the optimization with stochastic gradient descent, using the RMSProp method with a base learning rate of 0.001 and mini-batches of size 32.We apply Early stopping with patience 20,we wanted to monitor the validation loss at each epoch and after the validation loss has not improved after 20 epochs, training is interrupted.

\section{Result}
\subsection{Comparison with State-of-the-Art on HKR Dataset }
In this section, we will represent the results of applying the research model and compare its performance with the other published models which used different datasets to achieve a state-of-the-art scientific comparison. (Bluche, and Puigcerver) \cite{bluche2017gated,puigcerver2017multidimensional}. We divided our dataset into four partitions: train, valid, Test1, and Test2. Testing our model and the other models are implemented via twice tests of datasets as shown in Table. 1. 
Our network is trained from scratch, because there is no pre-training model or any transfer from another dataset is used before. For training, we used a RmsProp optimizer \cite{tieleman2012lecture} with an initial learning rate of 0.001 and mini-batches of size 32. Early stopping is applied after 20 non-improving epochs. Dropout \cite{srivastava2014dropout} was applied after several convolutional layers of all the networks with a probability of 0.5, whereas the standard performance measures are used for all results presented: the character error rate (CER) and word error rate (WER)\cite{frinken2014continuous}. The CER is determined as the distance from Levenshtein, which is the sum of the character substitution (S), insertion (I) and deletions (D) required to turn one string into the other, divided by the total number of characters in the ground truth word (N).
\begin{equation}
\begin{split}
CER = \frac{S+I+D}{N}	
\end{split}
\end{equation}
Similarly, the WER is calculated as the sum of  number of the term substitutions (Sw), insertion (Iw) and deletions (Dw),  which is necessary for the transformation of one string into the other, are divided by the total number of ground-truth terms (Nw). 
Our model was trained to minimize the validation loss value of the CTC function, training and validation Loss shown in Fig. 5.
\begin{equation}
\begin{split}
WER = \frac{Sw+Iw+Dw}{Nw}	
\end{split}
\end{equation}
\begin{figure}
\centerline{\includegraphics[width=\textwidth,keepaspectratio]{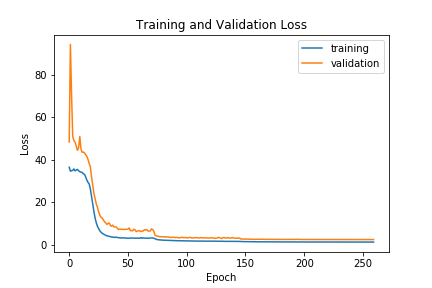}}

\caption{Training and validation loss.}
\end{figure}

\begin{table}
\caption{ Comparison of results on HKR dataset to previous methods.}
\begin{center}
\begin{tabular}{|c|c|c|c|c|c|c|c|}
    \hline
    \multirow{2}{*}{Model}& \multicolumn{3}{|c|}{ \textbf{Test1} }& \multicolumn{3}{|c|}{\textbf{Test2}} & \multirow{2}{*}{Params}\\
    \cline{2-7}
             &CER       & WER & SER & CER & WER & SER & \\
\hline
    \textbf{Our Model} & \textbf{0.045} & \textbf{0.192} & \textbf{0.253} & \textbf{0.064} & \textbf{0.24} & \textbf{0.361} & \textbf{885K} \\
    \hline
    Bluche & 0.161 & 0.596 & 0.673 & 0.101 & 0.374 & 0.510 & 728K\\
    \hline
    Puigcerver & 0.434 & 0.768  & 0.968 & 0.547 & 0.829 & 0.981 & 9.6M\\
    \hline
\end{tabular}
\end{center}
\end{table}

Neural networks use randomness by design to ensure that the function being approximated for the problem is effectively learned. Randomness is used because it can provide better performance with this type of machine learning algorithm than the others. Random initialization of the network weights is the most common type of randomness used in neural networks. While randomness can be used in other areas, here's some examples:
\begin{itemize}
    \item Initialization Randomness, such as weights. 
    \item Regularisation randomness, such as dropout. 
    \item Randomness in layers, like embedding of words. 
    \item Optimisation randomness, such as stochastic optimization.

\end{itemize}

\begingroup
\begin{table}[h!]
\caption{ CER, WER, SER for Test1, and Test2 for 10-time experiments.}
\setlength{\tabcolsep}{3pt}
\begin{center}
\begin{tabular}{|c|c|c|c|c|c|c|c|c|}
    \hline
    \multirow{2}{*}{Exp }& \multicolumn{3}{|c|}{ \textbf{Test1} }& \multicolumn{3}{|c|}{\textbf{Test2}} & \multirow{2}{*}{Seed}& \multirow{2}{*}{Time} \\
    \cline{2-7}
             &CER       & WER & SER & CER & WER & SER & &\\
    \hline
    1 & 0.045 & 0.192 & 0.253 & 0.064 & 0.245 & 0.361 & 1234 & 3 days, 2:52:28\\
    \hline
    2 & 0.046 & 0.209 & 0.276 & 0.063 & 0.245 & 0.363 & 50 & 2 days, 22:22:27 \\
    \hline
    3 & 0.039 & 0.178 & 0.243 & 0.065 & 0.250 & 0.367 & 70 & 4 days, 14:32:49\\
    \hline
    4 & 0.042 & 0.198 & 0.250 & 0.062 & 0.243 & 0.360 & 1225 & 7 days, 1:46:26\\
    \hline
    5 & 0.039 & 0.197 & 0.320 & 0.058 & 0.225 & 0.337 & 80 & 5 days, 0:13:21\\
    \hline
    6 & 0.042 & 0.194 & 0.258 & 0.062 & 0.246 & 0.364 & 500 & 3 days, 12:50:51\\
    \hline
    7 & 0.034 & 0.155 & 0.218  & 0.052 & 0.213  & 0.320 & 2000 & 2 days, 10:50:32\\
    \hline
    8 & 0.044 & 0.186 & 0.250 & 0.056 & 0.224 & 0.336 & 1334 & 2 days, 22:36:42\\
    
    \hline
    9 & 0.043 & 0.198 & 0.259 & 0.062 & 0.244 & 0.362 & 800 & 4 days, 6:15:17\\
    \hline
    10 &  0.039 & 0.184 & 0.245 & 0.059 & 0.234 & 0.346 & 150 & 5 days, 4:52:44\\
    \hline
    mean &  0.0413 & 0.1891 & 0.2572 &  0.0603 & 0.2369 & 0.3516 & - &- \\
    \hline
\end{tabular}
\end{center}
\end{table}
\endgroup

\subsection{Comparison with Other Datasets}
The results of our research model will be demonstrated and compared with other public datasets such as IAM\cite{marti2002iam}, Saint Gall\cite{fischer2011transcription},Bentham\cite{gatos2014ground}, and Washington\cite{fischer2012lexicon}. 
The IAM Handwriting Database 3.0 contains: 1539 pages of scanned text, 5685 isolated and labeled sentences, 657 authors contributed samples of their handwriting, 13353 isolated and labeled text lines and 115,320 isolated and labeled words. This dataset includes training, validation and test splits, where an author contributing to a training set can not occur in the validation or test splitting.
The presented Saint Gall database contains a handwritten historical manuscript with the following features: 9th century, Latin language, single writer, Carolingian script and parchment ink. Database Saint Gall contains:
60 pages, 1,410 text lines, 11,597 words, 4,890 word labels, 5,436 word spellings and 49 letters.
The Washington database was developed at the Library of Congress from George Washington Papers and has the following characteristics: eighteenth century, English language, two authors, longhand script and on paper ink. The Washington database includes: 20 pages, 656 lines of text, 4,894 instances of word, 1,471 classes of words and 82 letters.

Bentham 's writings contain a significant number of articles written by renowned British philosopher and reformist Jeremy Bentham (1748-1832). Currently, this sequence is transcribed by an amateur volunteer involved in the award-winning crowd-sourced initiative, Transcribe Bentham.
Now more than 6,000 documents have been transcribed through this public online site. Bentham data set is a subset of documents transcribed using TranScriptorium. This dataset is free and available in two parts for research purposes: the images and the GT.The GT provide information on the layout and transcription of each image on the line stage in PAGE format. Both sections have to be downloaded separately. In each section a comprehensive explanation is given of how the dataset is structured.We obtained state-of-the-art results on IAM, Saint Gall, Bentham and Washington databases shown in table. 3.
\begingroup
\begin{table}
\setlength{\tabcolsep}{3pt}
\caption{ Comparison of results on public datasets.}
\begin{center}
\begin{tabular}{|c|c|c|c|c|c|c|c|c|}
    \hline
    \multirow{2}{*}{Model}& \multicolumn{2}{|c|}{ \textbf{IAM} }& \multicolumn{2}{|c|}{\textbf{Saint Gall}} & \multicolumn{2}{|c|}{ \textbf{Washington} }& \multicolumn{2}{|c|}{ \textbf{Bentham} }\\
    \cline{2-9}
             &CER       & WER  & CER & WER  & CER & WER &CER & WER \\
\hline
    \textbf{Our Model} & \textbf{0.078} & \textbf{0.255} & \textbf{0.092} & \textbf{0.410} & \textbf{0.087} & \textbf{0.275} & \textbf{0.071} & \textbf{0.209} \\
    \hline
    Bluche & 0.107 & 0.318 & 0.101 & 0.439 & 0.215 &0.499& 0.113& 0.305 \\
    \hline
    Puigcerver &  0.082 & 0.270  & 0.145 & 0.569 & 0.286 & 0.556 &0.072 & 0.203 \\

    \hline
\end{tabular}
\end{center}
\end{table}
\endgroup

\subsection{Experiments}
The proposed and tested models have all been implemented using the Tensorflow
Both the proposed and the tested models have been implemented using the Tensorflow library \cite{abadi2016tensorflow} for Python, which allows for transparent use of highly optimized mathematical operations on GPU’s through Python. A computational graph is defined in the Python script, to define all operations that are necessary for the specific computations. Then the tensors evaluated and Tensorflow runs the necessary part of the computational graph implemented in efficient C code on the CPU, or on a GPU if any is made available to the script and the operations have a supported GPU implementation. 
Despite Tensorflow supports the use of multiple GPU’s, in our project implementation we utilized only a single GPU for each test run to make the processing easier. The experiments were run on a machine with 2x “Intel(R) Xeon(R) E-5-2680” CPU’s and 4x "NVIDIA Tesla k20x"
\subsection{Discussion}
The primary goal of this research work was to investigate and quantitatively compare the state-of-the-art RNN models to choose the best performing one in a handwritten Cyrillic postal- address recognition task. This goal also incorporates all efforts put on improving the best performing RNN model. According to experiment results, Our model demonstrated comparatively better results in terms of generalization and overall accuracy (see Table 1). since the dataset includes a small number of Kazakh language handwritings, the language characters have lower frequencies (distribution in the dataset) compared to other Cyrillic letters. Consequently, the above-mentioned models struggle to recognize these characters resulting in very low recognition rates. Hence, this affects the overall average CAR rates.
The dataset also includes non-alphabetic characters (such as ”. , !” and so on) with small distributions. Puigcerver and Bluche models seemed to prone to overfitting while being trained to Cyrillic handwriting. It seems that enriching the dataset with a variety of Kazakh and Russian words, and making it balanced will solve this issue.

In the proposed architecture, an "image-level," consisting of convolutions and a language-level model of recurrent layers, has been conceptually separated.  we trained our model for more than 10 times and all results are recorded in Table. 2. for each experiment trained, evaluated, and tested in the different random seed. On the other hand, Table. 4. shows training and validation losses in each experiment.
From table. 2. the time is different because we trained 4 experiments at the same time.

By training the encoder on a large amount of data, in the Russian and Kazakh languages and from various collections. We plan in future to use our model for other applications, including speech recognition, image tagging, video captioning, sign language translation, music composition, and genome sequencing, which may benefit from our approach. For example, a recurrent neural network transforms raw voice into character streams using the DeepSpeech speech recognition technique. Both streams of characters use CTC for logical words in text streams.
\begin{table}
\caption{ Training and Validation loss for 10-time experiments.}
\begin{center}
\begin{tabular}{|c|c|c|c|c|}
    \hline
    Exp & training loss & validation loss & Total epochs & Best epoch \\
    \hline
    1 & 1.21795921 & 2.34796703  & 260 &240 \\
    \hline
    2 & 1.26175002 & 2.43757006 & 246 &226 \\
    \hline
    3 &  1.44539043 & 2.44581821 & 277 &257\\
    \hline
    4 & 1.43913736 & 2.76414666 & 359 &339\\
    \hline
     5 &  1.28837166 & 2.37435660 & 250 &220\\
    \hline
    6 &  1.30836655 &  2.41045173 & 261 & 245\\
    \hline
    7 & 1.02203735  & 2.08543044  & 210 & 190\\
    \hline
    8 &  1.26351223 & 2.22102897  & 251 & 231\\
    \hline
    9 &  1.30336783 &  2.401427975 & 286 & 235\\
    \hline
    10 & 1.18126260 & 2.26886813  & 262  & 242\\
    \hline
    mean & 1.273115524 & 2.3757065805  &  - &- \\
    \hline
\end{tabular}
\end{center}
\end{table}

\section{Conclusion}
In this paper, we have presented attention encoder-decoder neural network architecture to achieve state-of-the-art results for Kazakh and Russian handwriting recognition. It is made of a Fully Gated convolutional encoder that extracts generic features of handwritten text, Attention mechanism, Bi-GRU decoder and CTC model to predict the sequence of characters. An important aspect is the attention mechanism that increases the capacity of the network to collect the most important information for every part of the output sequence also gated layers implemented in the encoder, which able to select the most important features and inhibit the others.

\section*{Acknowledgments}

This work was done with the support of grant funding for scientific projects of the MES RK No AR05135175 “Development and implementation of a system for recognizing handwritten addresses of written correspondence JSC “KazPost” using machine learning”.


\begin{thebibliography}{10}
\expandafter\ifx\csname url\endcsname\relax
  \def\url#1{\texttt{#1}}\fi
\expandafter\ifx\csname urlprefix\endcsname\relax\def\urlprefix{URL }\fi
\expandafter\ifx\csname href\endcsname\relax
  \def\href#1#2{#2} \def\path#1{#1}\fi

\bibitem{BUNKE19951399}
H.~Bunke, M.~Roth, E.~Schukat-Talamazzini, Off-line cursive handwriting
  recognition using hidden markov models, Pattern Recognition 28~(9) (1995)
  1399 -- 1413.
\newblock \href {https://doi.org/https://doi.org/10.1016/0031-3203(95)00013-P}
  {\path{doi:https://doi.org/10.1016/0031-3203(95)00013-P}}.

\bibitem{inproceedings}
A.~Toselli, E.~Vidal, Handwritten text recognition results on the bentham
  collection with improved classical n-gram-hmm methods, 2015, pp. 15--22.
\newblock \href {https://doi.org/10.1145/2809544.2809551}
  {\path{doi:10.1145/2809544.2809551}}.

\bibitem{rabiner1986introduction}
L.~Rabiner, B.~Juang, An introduction to hidden markov models, ieee assp
  magazine 3~(1) (1986) 4--16.

\bibitem{poritz1988hidden}
A.~B. Poritz, Hidden markov models: A guided tour, in: Proceedings of the IEEE
  Conference on Acoustics, Speech and Signal Processing (ICASSP), 1988, pp.
  7--13.

\bibitem{bahl1983maximum}
L.~R. Bahl, F.~Jelinek, R.~L. Mercer, A maximum likelihood approach to
  continuous speech recognition, IEEE transactions on pattern analysis and
  machine intelligence~(2) (1983) 179--190.

\bibitem{rabiner1989tutorial}
L.~R. Rabiner, A tutorial on hidden markov models and selected applications in
  speech recognition, Proceedings of the IEEE 77~(2) (1989) 257--286.

\bibitem{lee1990speech}
K.-F. Lee, H.-W. Hon, M.-Y. Hwang, X.~Huang, Speech recognition using hidden
  markov models: a cmu perspective, Speech Communication 9~(5-6) (1990)
  497--508.

\bibitem{gillies1992cursive}
A.~M. Gillies, Cursive word recognition using hidden markov models, in: Proc.
  Fifth US Postal Service Advanced Technology Conference, 1992, pp. 557--562.

\bibitem{caesar1993recognition}
T.~Caesar, J.~Gloger, A.~Kaltenmeier, E.~Mandler, Recognition of handwritten
  word images by statistical methods, in: Proceedings Int. Workshop on
  Frontiers in Handwriting Recognition, 1993, pp. 409--416.

\bibitem{mohamed1996handwritten}
M.~Mohamed, P.~Gader, Handwritten word recognition using segmentation-free
  hidden markov modeling and segmentation-based dynamic programming techniques,
  IEEE transactions on pattern analysis and machine intelligence 18~(5) (1996)
  548--554.

\bibitem{chen1994off}
M.-Y. Chen, A.~Kundu, J.~Zhou, Off-line handwritten word recognition using a
  hidden markov model type stochastic network, IEEE transactions on Pattern
  analysis and Machine Intelligence 16~(5) (1994) 481--496.

\bibitem{gilloux1995strategies}
M.~Gilloux, M.~Leroux, J.-M. Bertille, Strategies for cursive script
  recognition using hidden markov models, Machine Vision and Applications 8~(4)
  (1995) 197--205.

\bibitem{NIPS2007_3213}
A.~Graves, M.~Liwicki, H.~Bunke, J.~Schmidhuber, S.~Fern\'{a}ndez,
  Unconstrained on-line handwriting recognition with recurrent neural networks,
  in: J.~C. Platt, D.~Koller, Y.~Singer, S.~T. Roweis (Eds.), Advances in
  Neural Information Processing Systems 20, Curran Associates, Inc., 2008, pp.
  577--584.

\bibitem{7814068}
P.~{Voigtlaender}, P.~{Doetsch}, H.~{Ney}, Handwriting recognition with large
  multidimensional long short-term memory recurrent neural networks, in: 2016
  15th International Conference on Frontiers in Handwriting Recognition
  (ICFHR), 2016, pp. 228--233.

\bibitem{4531750}
A.~{Graves}, M.~{Liwicki}, S.~{Fernández}, R.~{Bertolami}, H.~{Bunke},
  J.~{Schmidhuber}, A novel connectionist system for unconstrained handwriting
  recognition, IEEE Transactions on Pattern Analysis and Machine Intelligence
  31~(5) (2009) 855--868.

\bibitem{xu2015show}
K.~Xu, J.~Ba, R.~Kiros, K.~Cho, A.~Courville, R.~Salakhudinov, R.~Zemel,
  Y.~Bengio, Show, attend and tell: Neural image caption generation with visual
  attention, in: International conference on machine learning, 2015, pp.
  2048--2057.

\bibitem{bahdanau2014neural}
D.~Bahdanau, K.~Cho, Y.~Bengio, Neural machine translation by jointly learning
  to align and translate, arXiv preprint arXiv:1409.0473 (2014).

\bibitem{abdallah2020automated}
A.~Abdallah, M.~Kasem, M.~Hamada, S.~Sdeek, Automated question answer medical
  model based on deep learning technology, arXiv preprint arXiv:2005.10416
  (2020).

\bibitem{huang2019attention}
L.~Huang, W.~Wang, J.~Chen, X.-Y. Wei, Attention on attention for image
  captioning, in: Proceedings of the IEEE International Conference on Computer
  Vision, 2019, pp. 4634--4643.

\bibitem{hamada2019sentimental}
M.~A. Hamada, K.~Sultanbek, B.~Alzhanov, B.~Tokbanov, Sentimental text
  processing tool for russian language based on machine learning algorithms,
  in: Proceedings of the 5th International Conference on Engineering and MIS,
  2019, pp. 1--6.

\bibitem{chorowski2015attention}
J.~K. Chorowski, D.~Bahdanau, D.~Serdyuk, K.~Cho, Y.~Bengio, Attention-based
  models for speech recognition, in: Advances in neural information processing
  systems, 2015, pp. 577--585.

\bibitem{bluche2017scan}
T.~Bluche, J.~Louradour, R.~Messina, Scan, attend and read: End-to-end
  handwritten paragraph recognition with mdlstm attention, in: 2017 14th IAPR
  International Conference on Document Analysis and Recognition (ICDAR),
  Vol.~1, IEEE, 2017, pp. 1050--1055.

\bibitem{kang2018convolve}
L.~Kang, J.~I. Toledo, P.~Riba, M.~Villegas, A.~Forn{\'e}s, M.~Rusinol,
  Convolve, attend and spell: An attention-based sequence-to-sequence model for
  handwritten word recognition, in: German Conference on Pattern Recognition,
  Springer, 2018, pp. 459--472.

\bibitem{cho2014learning}
K.~Cho, B.~Van~Merri{\"e}nboer, C.~Gulcehre, D.~Bahdanau, F.~Bougares,
  H.~Schwenk, Y.~Bengio, Learning phrase representations using rnn
  encoder-decoder for statistical machine translation, arXiv preprint
  arXiv:1406.1078 (2014).

\bibitem{nurseitov2020hkr}
D.~Nurseitov, K.~Bostanbekov, D.~Kurmankhojayev, A.~Alimova, A.~Abdallah, Hkr
  for handwritten kazakh \& russian database, arXiv preprint arXiv:2007.03579
  (2020).

\bibitem{gupta2011offline}
A.~Gupta, M.~Srivastava, C.~Mahanta, Offline handwritten character recognition
  using neural network, in: 2011 IEEE International Conference on Computer
  Applications and Industrial Electronics (ICCAIE), IEEE, 2011, pp. 102--107.

\bibitem{bianne2011dynamic}
A.-L. Bianne-Bernard, F.~Menasri, R.~A.-H. Mohamad, C.~Mokbel, C.~Kermorvant,
  L.~Likforman-Sulem, Dynamic and contextual information in hmm modeling for
  handwritten word recognition, IEEE transactions on pattern analysis and
  machine intelligence 33~(10) (2011) 2066--2080.

\bibitem{bluche2017gated}
T.~Bluche, R.~Messina, Gated convolutional recurrent neural networks for
  multilingual handwriting recognition, in: 2017 14th IAPR International
  Conference on Document Analysis and Recognition (ICDAR), Vol.~1, IEEE, 2017,
  pp. 646--651.

\bibitem{puigcerver2017multidimensional}
J.~Puigcerver, Are multidimensional recurrent layers really necessary for
  handwritten text recognition?, in: 2017 14th IAPR International Conference on
  Document Analysis and Recognition (ICDAR), Vol.~1, IEEE, 2017, pp. 67--72.

\bibitem{wang2020writer}
Z.-R. Wang, J.~Du, J.-M. Wang, Writer-aware cnn for parsimonious hmm-based
  offline handwritten chinese text recognition, Pattern Recognition 100 (2020)
  107102.

\bibitem{graves2006connectionist}
A.~Graves, S.~Fern{\'a}ndez, F.~Gomez, J.~Schmidhuber, Connectionist temporal
  classification: labelling unsegmented sequence data with recurrent neural
  networks, in: Proceedings of the 23rd international conference on Machine
  learning, 2006, pp. 369--376.

\bibitem{srivastava2014dropout}
N.~Srivastava, G.~Hinton, A.~Krizhevsky, I.~Sutskever, R.~Salakhutdinov,
  Dropout: a simple way to prevent neural networks from overfitting, The
  journal of machine learning research 15~(1) (2014) 1929--1958.

\bibitem{tieleman2012lecture}
T.~Tieleman, G.~Hinton, Lecture 6.5-rmsprop: Divide the gradient by a running
  average of its recent magnitude, COURSERA: Neural networks for machine
  learning 4~(2) (2012) 26--31.

\bibitem{frinken2014continuous}
V.~Frinken, H.~Bunke, Continuous handwritten script recognition. (2014).

\bibitem{marti2002iam}
U.-V. Marti, H.~Bunke, The iam-database: an english sentence database for
  offline handwriting recognition, International Journal on Document Analysis
  and Recognition 5~(1) (2002) 39--46.

\bibitem{fischer2011transcription}
A.~Fischer, V.~Frinken, A.~Forn{\'e}s, H.~Bunke, Transcription alignment of
  latin manuscripts using hidden markov models, in: Proceedings of the 2011
  Workshop on Historical Document Imaging and Processing, 2011, pp. 29--36.

\bibitem{gatos2014ground}
B.~Gatos, G.~Louloudis, T.~Causer, K.~Grint, V.~Romero, J.~A. S{\'a}nchez,
  A.~H. Toselli, E.~Vidal, Ground-truth production in the transcriptorium
  project, in: 2014 11th IAPR International Workshop on Document Analysis
  Systems, IEEE, 2014, pp. 237--241.

\bibitem{fischer2012lexicon}
A.~Fischer, A.~Keller, V.~Frinken, H.~Bunke, Lexicon-free handwritten word
  spotting using character hmms, Pattern Recognition Letters 33~(7) (2012)
  934--942.

\bibitem{abadi2016tensorflow}
M.~Abadi, A.~Agarwal, P.~Barham, E.~Brevdo, Z.~Chen, C.~Citro, G.~S. Corrado,
  A.~Davis, J.~Dean, M.~Devin, et~al., Tensorflow: Large-scale machine learning
  on heterogeneous distributed systems, arXiv preprint arXiv:1603.04467 (2016).

\end{thebibliography}
\end{document}